\documentclass{article}
\usepackage{times}
\usepackage{graphicx} % more modern
\usepackage{natbib}

\usepackage{algorithm}
\usepackage{algorithmic}

\usepackage{hyperref}

\usepackage{booktabs}
\usepackage{blindtext}
\usepackage{graphicx}
\usepackage{caption,subcaption}
\usepackage{etoolbox}
\usepackage{bm}
\usepackage{scrextend}
\usepackage[flushleft]{threeparttable}
\usepackage[usenames,dvipsnames,svgnames,table]{xcolor}
\usepackage[normalem]{ulem}
\usepackage{amsmath,amsthm, amssymb, latexsym}
\usepackage{epsfig}
\usepackage{multirow}
\usepackage{pifont}
\usepackage{bigints}
\usepackage{soul}
\usepackage{marginnote}
\usepackage{colortbl}
\usepackage{subcaption}

\newcommand{\mean}{\mathbb{E}}%{{\rm I\kern-.3em E}}
\newcommand{\var}{{\rm I\kern-.3em D}}

\newcommand{\svec}[1]{\mathbf{#1}}
\newcommand{\RR}{\mathbb{R}}
\newcommand{\cond}{\,|\,}
\newcommand{\Normal}{\mathcal{N}}
\newcommand{\diag}{\mathrm{diag}}

\usepackage[accepted]{icml2017} 

\icmltitlerunning{Variational Dropout Sparsifies Deep Neural Networks}

\begin{document} 

\twocolumn[
\icmltitle{Variational Dropout Sparsifies Deep Neural Networks}

\icmlsetsymbol{equal}{*}

\begin{icmlauthorlist}
\icmlauthor{Dmitry Molchanov}{ya,sk,equal}
\icmlauthor{Arsenii Ashukha}{hse,mipt,equal}
\icmlauthor{Dmitry Vetrov}{hse,ya}
\end{icmlauthorlist}

\icmlaffiliation{sk}{Skolkovo Institute of Science and Technology, Skolkovo Innovation Center, Moscow, Russia}
\icmlaffiliation{mipt}{Moscow Institute of Physics and Technology, Moscow, Russia}
\icmlaffiliation{hse}{National Research University Higher School of Economics, Moscow, Russia}
\icmlaffiliation{ya}{Yandex, Russia}

\icmlcorrespondingauthor{Dmitry Molchanov}{dmitry.molchanov@skolkovotech.ru}
\icmlcorrespondingauthor{Arsenii Ashukha}{ars.ashuha@gmail.com}
\icmlcorrespondingauthor{Dmitry Vetrov}{vetrovd@yandex.ru}

\icmlkeywords{}

\vskip 0.3in
]



\printAffiliationsAndNotice{\icmlEqualContribution} % otherwise use the standard text.
%\footnotetext{hi}

\begin{abstract} 
We explore a recently proposed Variational Dropout technique that provided an elegant Bayesian interpretation to Gaussian Dropout.
We extend Variational Dropout to the case when dropout rates are unbounded, propose a way to reduce the variance of the gradient estimator and report first experimental results with individual dropout rates per weight.
Interestingly, it leads to extremely sparse solutions both in fully-connected and convolutional layers.
This effect is similar to automatic relevance determination effect in empirical Bayes but has a number of advantages.
We reduce the number of parameters up to $280$ times on LeNet architectures and up to $68$ times on VGG-like networks with a negligible decrease of accuracy.
\end{abstract}

\section{Introduction}

Deep neural networks (DNNs) are a widely popular family of models which is currently state-of-the-art in many important problems \cite{szegedy2016inception, silver2016mastering}.
However, DNNs often have many more parameters than the number of the training instances.
This makes them prone to overfitting \cite{hinton2012improving,zhang2016understanding} and necessitates using regularization.
A commonly used regularizer is Binary Dropout \cite{hinton2012improving} that prevents co-adaptation of neurons by randomly dropping them during training.
An equally effective alternative is Gaussian Dropout \cite{srivastava2014dropout} that multiplies the outputs of the neurons by Gaussian random noise.

Dropout requires specifying the dropout rates which are the probabilities of dropping a neuron.
The dropout rates are typically optimized using grid search.
To avoid the exponential complexity of optimizing multiple hyperparameters, the dropout rates are usually shared for all layers.
Recently it was shown that dropout can be seen as a special case of Bayesian regularization \cite{gal2015dropout,kingma2015vdo}.
It is an important theoretical result that justifies dropout and at the same time allows us to tune individual dropout rates for each weight, neuron or layer in a Bayesian way. 

Instead of injecting noise we can regularize a model by reducing the number of its parameters.
This technique is especially attractive in the case of deep neural networks.
Modern neural networks contain hundreds of millions of parameters \cite{szegedy2015going, he2015deep} and require a lot of computational and memory resources. 
It restricts us from using deep neural networks when those resources are limited.
Inducing sparsity during training of DNNs leads to regularization, compression, and acceleration of the resulting model \cite{han2015deep,scardapane2016group}.

Sparse Bayesian Learning \cite{tipping2001sparse} provides a principled framework for training of sparse models without the manual tuning of hyperparameters.
Unfortunately, this approach does not extend straightforwardly to DNNs.
During past several years, a number of papers \cite{hoffman2013stochastic, kingma2013auto, rezende2014stochastic} on scalable variational inference have appeared.
These techniques make it possible to train Bayesian Deep Neural Networks using stochastic optimization and provide us an opportunity to transfer Bayesian regularization techniques from simple models to DNNs.

In this paper, we study Variational Dropout \cite{kingma2015vdo} in the case when each weight of a model has its individual dropout rate.
We propose Sparse Variational Dropout that extends Variational Dropout to all possible values of dropout rates and leads to a sparse solution.
To achieve this goal, we provide a new approximation of the KL-divergence term in Variational Dropout objective that is tight on the full domain.
We also propose a way to greatly reduce the variance of the stochastic gradient estimator and show that it leads to a much faster convergence and a better value of the objective function.
We show theoretically that Sparse Variational Dropout applied to linear models can lead to a sparse solution.
Like classical Sparse Bayesian models, our method provides the Automatic Relevance Determination effect, but overcomes certain disadvantages of empirical Bayes.

Our experiments show that Sparse Variational Dropout leads to a high level of sparsity in fully-connected and convolutional layers of Deep Neural Networks.
Our method achieves a state-of-the-art sparsity level on LeNet architectures and scales on larger networks like VGG with negligible performance drop.
Also we show that our method fails to overfit on randomly labeled data unlike Binary Dropout networks.

\section{Related Work}
\label{sec:related}

Deep Neural Nets are prone to overfitting and regularization is used to address this problem.
Several successful techniques have been proposed for DNN regularization, among them are Dropout \cite{srivastava2014dropout}, DropConnect \cite{wan2013regularization}, Max Norm Constraint \cite{srivastava2014dropout}, Batch Normalization \cite{ioffe2015batch}, etc.

Another way to regularize deep model is to reduce the number of parameters.
One possible approach is to use tensor decompositions \cite{novikov2015tensorizing, garipov2016ultimate}.
Another approach is to induce sparsity into weight matrices.
Most recent works on sparse neural networks use pruning \cite{han2015learning}, elastic net regularization \cite{lebedev2015fast, liu2015sparse, scardapane2016group, wen2016lssdnn} or composite techniques \cite{han2015deep,guo2016dynamic,ullrich2017soft}.

Sparsity can also be obtained by using the Sparse Bayesian Learning framework \cite{tipping2001sparse}. 
Automatic Relevance Determination was introduced in \cite{neal2012bayesian, mackay1994bayesian}, where small neural networks were trained with ARD regularization on the input layer.
This approach was later studied on linear models like the Relevance Vector Machine \cite{tipping2001sparse} and other kernel methods \cite{van2001automatic}.
In the Relevance Tagging Machine model \cite{molchanov2015relevance} Beta prior distribution is used to obtain the ARD effect in a similar setting.

Recent works on Bayesian DNNs \cite{kingma2013auto, rezende2014stochastic, scardapane2016group} provide different ways to train deep models with a huge number of parameters in a Bayesian way.
These techniques can be applied to improve latent variables models \cite{kingma2013auto}, to prevent overfitting and to obtain model uncertainty \cite{gal2015dropout}.
Recently several works on efficient training of Sparse Bayesian Models have appeared \cite{challis2013gkl,titsias2014dsvi}.
Soft Weights Sharing \cite{ullrich2017soft} uses an approach, similar to Sparse Bayesian Learning framework, to obtain a sparse and quantized Bayesian Deep Neural Network, but utilizes a more flexible family of prior distributions.

Variational Dropout \cite{kingma2015vdo} is an elegant interpretation of Gaussian Dropout as a special case of Bayesian regularization.
This technique allows us to tune dropout rate and can, in theory, be used to set individual dropout rates for each layer, neuron or even weight.
However, that paper uses a limited family for posterior approximation that does not allow for ARD effect.
Other Bayesian interpretations of dropout training have also appeared during past several years \cite{maeda2014bayesian,gal2015dropout,srinivas2016generalized}.
Generalized Dropout \cite{srinivas2016generalized} provides a way to tune individual dropout rates for neurons, but uses a biased gradient estimator.
Also, the posterior distribution is modelled by a delta function, so the resulting neural network is effectively not Bayesian.
Variational Spike-and-Slab Neural Networks \cite{louizos2015smart} is yet another Bayesian interpretation of Binary Dropout that allows for tuning of individual dropout rates and also leads to a sparse solution.
Unfortunately, this procedure does not scale well with model width and depth.

\section{Preliminaries}
\label{sec:prelim}
We begin by describing the Bayesian Inference and Stochastic Variational Inference frameworks.
Then we describe Variational Dropout, a recently proposed Bayesian regularization technique \cite{kingma2015vdo}.
\subsection{Bayesian Inference}
Consider a dataset $\mathcal{D}$ which is constructed from $N$ pairs of objects $(x_n, y_n)_{n=1}^N$.
Our goal is to tune the parameters $w$ of a model $p(y\cond x, w)$ that predicts $y$ given $x$ and $w$.
In Bayesian Learning we usually have some prior knowledge about weights $w$, which is expressed in terms of a prior distribution $p(w)$.
After data $\mathcal{D}$ arrives, this prior distribution is transformed into a posterior distribution $p(w\cond \mathcal{D})=p(\mathcal{D}\cond w)p(w)/p(\mathcal{D})$.
This process is called \emph{Bayesian Inference}.
Computing posterior distribution using the Bayes rule usually involves computation of intractable multidimensional integrals, so we need to use approximation techniques.

One of such techniques is \emph{Variational Inference}.
In this approach the posterior distribution $p(w\cond\mathcal{D})$ is approximated by a parametric distribution $q_\phi(w)$.
The quality of this approximation is measured in terms of the Kullback-Leibler divergence $D_{KL}(q_\phi(w)\,\|\,p(w\cond \mathcal{D}))$.
The optimal value of variational parameters $\phi$ can be found by maximization of the \emph{variational lower bound}:
\begin{equation}
   \mathcal{L}(\phi) =  L_\mathcal{D}(\phi) - D_{KL}(q_\phi(w)\,\|\,p(w)) \to\max_{\phi\in\Phi}
   \label{eq:elbo}
\end{equation}
\vspace{-1em}
\begin{equation}
   L_\mathcal{D}(\phi) = \sum_{n=1}^N \mean_{q_\phi(w)}[\log p(y_n\cond x_n, w)]
   \label{lD}
\end{equation}
It consists of two parts, the expected log-likelihood $L_\mathcal{D}(\phi)$ and the KL-divergence $D_{KL}(q_\phi(w)\,\|\,p(w))$, which acts as a regularization term.

\subsection{Stochastic Variational Inference}
\label{sec:vi}
In the case of complex models expectations in (\ref{eq:elbo}) and (\ref{lD}) are intractable.
Therefore the variational lower bound (\ref{eq:elbo}) and its gradients can not be computed exactly.
However, it is still possible to estimate them using sampling and optimize the variational lower bound using stochastic optimization.

We follow \cite{kingma2013auto} and use the Reparameterization Trick to obtain an unbiased differentiable minibatch-based Monte Carlo estimator of the expected log-likelihood (\ref{eq:sgvlb}).
The main idea is to represent the parametric noise $q_{\phi}(w)$ as a deterministic differentiable function $w = f(\phi, \epsilon)$ of a non-parametric noise $\epsilon \thicksim p(\epsilon)$. 
This trick allows us to obtain an unbiased estimate of $\nabla_{\phi} L_\mathcal{D}(q_\phi)$.
Here we denote objects from a mini-batch as $(\tilde{x}_m, \tilde{y}_m)_{m=1}^M$.
\begin{equation}
    \label{eq:sgvlb}
    \mathcal{L}(\phi)\!\simeq\!\mathcal{L}^{\text{\tiny \emph{SGVB}}}(\phi)\!=\!L_{\mathcal{D}}^{\text{\tiny \emph{SGVB}}}(\phi)-D_{KL}(q_\phi(w)\|p(w))
\end{equation}\vspace{-1.5em}
\begin{equation}
   L_\mathcal{D}(\phi)\!\simeq\!L_\mathcal{D}^{\text{\tiny \emph{SGVB}}}(\phi)\!=\!\frac{N}{M}\!\sum_{m = 1}^{M} \log p(\tilde{y}_m|\tilde{x}_m, f(\phi, \epsilon_m))
\end{equation}\vspace{-1.5em}
\begin{equation}
   \nabla_{\phi} L_\mathcal{D}(\phi) \!\simeq\!\frac{N}{M}\!\sum_{m = 1}^{M} \nabla_{\phi}\log p(\tilde{y}_m|\tilde{x}_m, f(\phi, \epsilon_m))
\end{equation}
The Local Reparameterization Trick is another technique that reduces the variance of this gradient estimator even further \cite{kingma2015vdo}.
The idea is to sample separate weight matrices for each data-point inside mini-batch.
It is computationally hard to do it straight-forwardly, but it can be done efficiently by moving the noise from weights to activations \cite{wang2013fast,kingma2015vdo}.

\subsection{Variational Dropout}
\label{sec:vdo}
In this section we consider a single fully-connected layer with $I$ input neurons and $O$ output neurons before a non-linearity.
We denote an output matrix as $B^{M\times O}$, input matrix as $A^{M\times I}$ and a weight matrix as ${W}^{I\times O}$.
We index the elements of these matrices as $b_{mj}$, $a_{mi}$ and $w_{ij}$ respectively.
Then $B=AW$.

Dropout is one of the most popular regularization methods for deep neural networks.
It injects a multiplicative random noise $\Xi$ to the layer input $A$ at each iteration of training procedure \cite{hinton2012improving}.
\begin{equation}
\label{eqdo}
  B = (A\odot \Xi) W,~\text{with}~~~\xi_{mi} \thicksim p(\xi)
\end{equation}
The original version of dropout, so-called Bernoulli or Binary Dropout, was presented with $\xi_{mi} \thicksim \mathrm{Bernoulli}(1-p)$ \cite{hinton2012improving}.
It means that each element of the input matrix is put to zero with probability $p$, also known as a \emph{dropout rate}.
Later the same authors reported that Gaussian Dropout with continuous noise $\xi_{mi} \thicksim \Normal(1, \alpha=\frac{p}{1-p})$ works as well and is similar to Binary Dropout with dropout rate $p$ \cite{srivastava2014dropout}.
It is beneficial to use continuous noise instead of discrete one because multiplying the inputs by a Gaussian noise is equivalent to putting Gaussian noise on the weights.
This procedure can be used to obtain a posterior distribution over the model's weights \cite{wang2013fast, kingma2015vdo}.
That is, putting multiplicative Gaussian noise $\xi_{ij}\sim \Normal(1, \alpha)$ on a weight $w_{ij}$ is equivalent to sampling of $w_{ij}$ from $q(w_{ij}\cond\theta_{ij},\alpha) = \Normal(w_{ij}\cond\theta_{ij}, \alpha\theta_{ij}^2)$.
Now $w_{ij}$ becomes a random variable parametrized by $\theta_{ij}$.
\begin{equation}
\label{eq:gdort}
\begin{gathered}
\!\!w_{ij}=\theta_{ij}{\xi}_{ij}=\theta_{ij}(1+\sqrt{\alpha}{\epsilon}_{ij})\sim \Normal(w_{ij}\cond\theta_{ij}, \alpha\theta_{ij}^2)\\
\!\!{\epsilon}_{ij} \thicksim \Normal(0, 1)
\end{gathered}
\end{equation}
Gaussian Dropout training is equivalent to stochastic optimization of the expected log likelihood (\ref{lD}) in the case when we use the reparameterization trick and draw a single sample $W\thicksim q(W\cond\theta,\alpha)$ per minibatch to estimate the expectation.
Variational Dropout extends this technique and explicitly uses $q(W\cond\theta,\alpha)$ as an approximate posterior distribution for a model with a special prior on the weights.
The parameters $\theta$ and $\alpha$ of the distribution $q(W\cond\theta,\alpha)$ are tuned via stochastic variational inference, i.e. $\phi=(\theta, \alpha)$ are the variational parameters, as denoted in Section~\ref{sec:vi}.
The prior distribution $p(W)$ is chosen to be improper log-scale uniform to make the Variational Dropout with fixed $\alpha$ equivalent to Gaussian Dropout \cite{kingma2015vdo}.
\begin{equation}
\label{eq:prior}
% \begin{gathered}
    p(\log|w_{ij}|) = \mathrm{const}\ \ \Leftrightarrow\ \ p(|w_{ij}|)\propto\frac{1}{|w_{ij}|}
% \end{gathered}
\end{equation}
In this model, it is the only prior distribution that makes variational inference consistent with Gaussian Dropout \cite{kingma2015vdo}.
When parameter $\alpha$ is fixed, the $D_{KL}(q(W\cond\theta,\alpha)\,\|\,p(W))$ term in the variational lower bound (\ref{eq:elbo}) does not depend on $\theta$ \cite{kingma2015vdo}.
Maximization of the variational lower bound (\ref{eq:elbo}) then becomes equivalent to maximization of the expected log-likelihood (\ref{lD}) with fixed parameter $\alpha$.
It means that Gaussian Dropout training is exactly equivalent to Variational Dropout with fixed $\alpha$.
However, Variational Dropout provides a way to train dropout rate $\alpha$ by optimizing the variational lower bound (\ref{eq:elbo}).
Interestingly, dropout rate $\alpha$ now becomes a variational parameter and not a hyperparameter.
In theory, it  allows us to train individual dropout rates $\alpha_{ij}$ for each layer, neuron or even weight \cite{kingma2015vdo}.
However, no experimental results concerning the training of individual dropout rates were reported in the original paper.
Also, the approximate posterior family was manually restricted to the case $\alpha\leq1$.

\section{Sparse Variational Dropout}
\label{sec:sparse}
In the original paper, authors reported difficulties in training the model with large values of dropout rates $\alpha$ \cite{kingma2015vdo} and only considered the case of $\alpha \leq 1$, which corresponds to a binary dropout rate $p\leq0.5$.
However, the case of large $\alpha_{ij}$ is very exciting (here we mean separate $\alpha_{ij}$ per weight).
High dropout rate $\alpha_{ij} \to +\infty$ corresponds to a binary dropout rate that approaches $p = 1$.
It effectively means that the corresponding weight or neuron is always ignored and can be removed from the model. 
In this work, we consider the case of individual $\alpha_{ij}$ for each weight of the model.

\subsection{Additive Noise Reparameterization}
Training Neural Networks with Variational Dropout is difficult when dropout rates $\alpha_{ij}$ are large because of a huge variance of stochastic gradients \cite{kingma2015vdo}.
The cause of large gradient variance arises from multiplicative noise.
To see it clearly, we can rewrite the gradient of $\mathcal{L}^{\text{\tiny \emph{SGVB}}}$ w.r.t. $\theta_{ij}$ as follows.
\begin{equation}    
\label{eq:dldtheta}
    \frac{\partial\mathcal{L}^{\text{\tiny \emph{SGVB}}}}{\partial\theta_{ij}}=
\frac{\partial\mathcal{L}^{\text{\tiny \emph{SGVB}}}}{\partial w_{ij}}\cdot\frac{\partial w_{ij}}{\partial\theta_{ij}}
\end{equation}
In the case of original parametrization $(\theta, \alpha)$ the second multiplier in (\ref{eq:dldtheta}) is very noisy if $\alpha_{ij}$ is large.
\begin{equation}
\label{eq:vdovar}
\begin{gathered}
w_{ij} = \theta_{ij} (1 + \sqrt{\alpha_{ij}} \cdot \epsilon_{ij}),\\
\frac{\partial w_{ij}}{\partial\theta_{ij}} = 1 + \sqrt{\alpha_{ij}} \cdot \epsilon_{ij},\\
\epsilon_{ij}\sim\Normal(0,1)
\end{gathered}
\end{equation}
We propose a trick that allows us to drastically reduce the variance of this term in the case when $\alpha_{ij}$ is large.
The idea is to replace the multiplicative noise term $1+\sqrt{\alpha_{ij}} \cdot\epsilon_{ij}$ with an exactly equivalent additive noise term $\sigma_{ij}\cdot\epsilon_{ij}$, where $\sigma_{ij}^2=\alpha_{ij}\theta_{ij}^2$ is treated as a new independent variable.
After this trick we will optimize the variational lower bound w.r.t. $(\theta, \sigma)$.
However, we will still use $\alpha$ throughout the paper, as it has a nice interpretation as a dropout rate.
\begin{equation}
\label{eq:oursparamvar}
\begin{gathered}
    w_{ij} = \theta_{ij} (1 + \sqrt{\alpha_{ij}}\cdot \epsilon_{ij}) = \theta_{ij} + \sigma_{ij} \cdot \epsilon_{ij}\\ \frac{\partial w_{ij}}{\partial\theta_{ij}} = 1,~~~~~~
    \epsilon_{ij} \sim \mathcal{N}(0, 1)
\end{gathered}
\end{equation}
From (\ref{eq:oursparamvar}) we can see that $\frac{\partial w_{ij}}{\partial\theta_{ij}}$ now has no injected noise, but the distribution over $w_{ij} \sim q(w_{ij}\cond\theta_{ij},\sigma_{ij}^2)$ remains exactly the same.
The objective function and the posterior approximating family are unaltered.
The only thing that changed is the parametrization of the approximate posterior.
However, the variance of a stochastic gradient is greatly reduced.
Using this trick, we avoid the problem of large gradient variance and can train the model within the full range of $\alpha_{ij}\in (0, +\infty)$.

It should be noted that the Local Reparametrization Trick does not depend on parametrization, so it can also be applied here to reduce the variance even further.
In our experiments, we use both Additive Noise Reparameterization and the Local Reparameterization Trick. 
We provide the final expressions for the outputs of fully-connected and convolutional layers for our model in Section~\ref{sec:formulae}.

\subsection{Approximation of the KL Divergence}

\begin{figure}[!tp]
    \centering
    \includegraphics[width=0.4\textwidth]{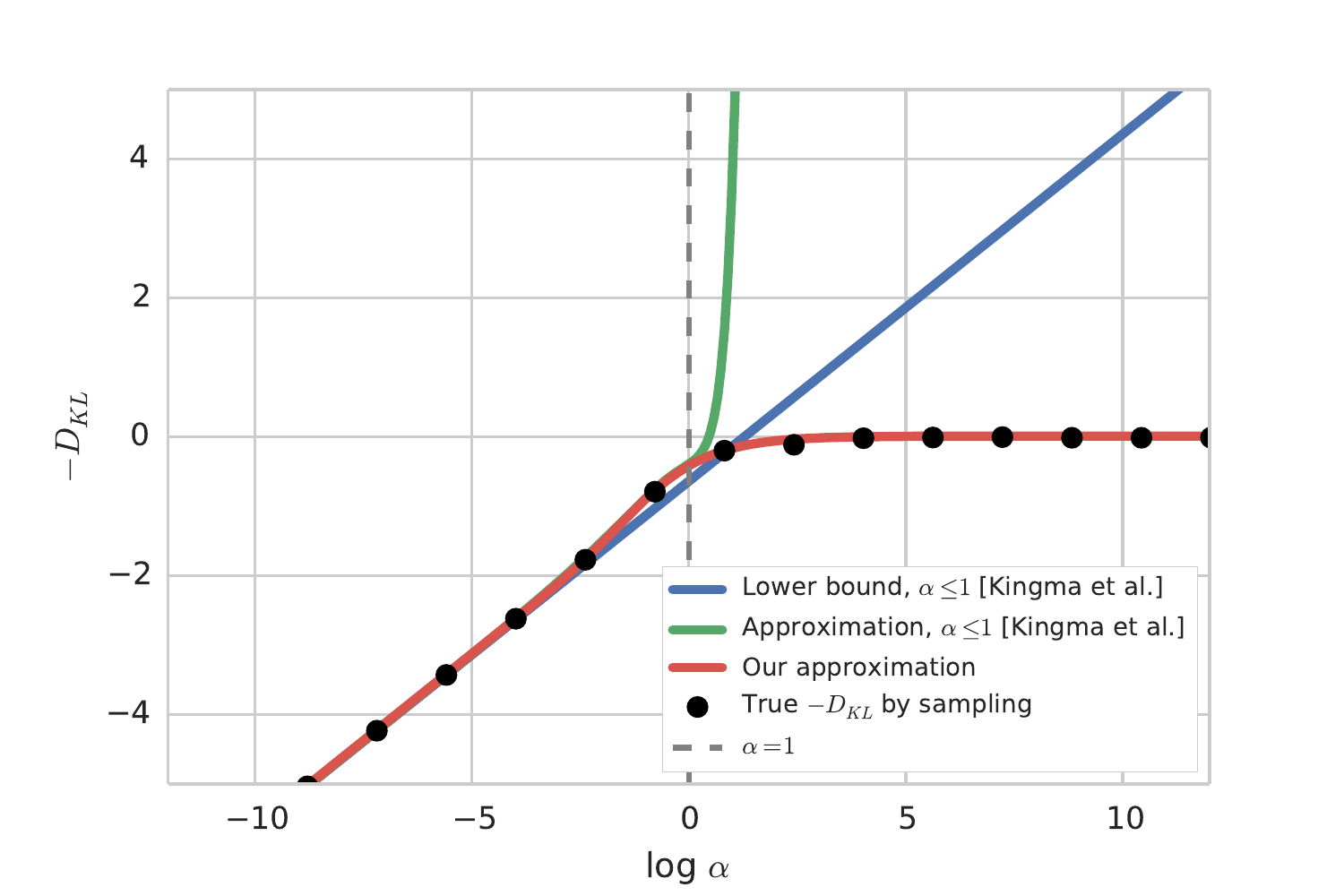}
    \caption{Different approximations of KL divergence: blue and green ones \cite{kingma2015vdo} are tight only for $\alpha \leq 1$; black one is the true value, estimated by sampling; red one is our approximation.}
    \label{fig:kl}
\end{figure}
As the prior and the approximate posterior are fully factorized, the full KL-divergence term in the lower bound (\ref{eq:elbo}) can be decomposed into a sum:
\begin{equation}
\begin{gathered}
D_{KL}(q(W\cond\theta,\alpha)\|\,p(W))=\\
= \sum_{ij} D_{KL}(q(w_{ij}\cond\theta_{ij}, \alpha_{ij})\,\|\,p(w_{ij})) 
\end{gathered}
\end{equation}
The log-scale uniform prior distribution is an improper prior, so the KL divergence can only be calculated up to an additive constant $\mathrm{C}$ \cite{kingma2015vdo}.
\begin{equation}
\begin{gathered}
    -D_{KL}(q(w_{ij}\cond\theta_{ij}, \alpha_{ij})\,\|\,p(w_{ij}))=
    \\
    =\frac12\log\alpha_{ij}-\mean_{\epsilon\sim\mathcal{N}(1, \alpha_{ij})}\log|\epsilon|+\mathrm{C}
    \label{eq:akl}
\end{gathered}
\end{equation}
In the Variational Dropout model this term is intractable, as the expectation $\mean_{\epsilon\sim\mathcal{N}(1, \alpha_{ij})}\log|\epsilon|$ in (\ref{eq:akl}) cannot be computed analytically \cite{kingma2015vdo}.
However, this term can be sampled and then approximated.
Two different approximations were provided in the original paper, however they are accurate only for small values of the dropout rate $\alpha$ ($\alpha\le 1$).
We propose another approximation (\ref{eq:KL}) that is tight for all values of alpha.
Here $\sigma(\cdot)$ denotes the sigmoid function.
Different approximations and the true value of $-D_{KL}$ are presented in Fig.~\ref{fig:kl}.
Original $-D_{KL}$ was obtained by averaging over $10^7$ samples of $\epsilon$ with less than $2\times10^{-3}$ variance of the estimation.
\begin{equation}
\begin{gathered}
    -D_{KL}(q(w_{ij}\cond\theta_{ij}, \alpha_{ij})\,\|\,p(w_{ij})) \approx
    \\
    \approx k_1\sigma(k_2 + k_3\log \alpha_{ij})) - 0.5\log(1+\alpha_{ij}^{-1}) + \mathrm{C}
    \label{eq:KL}\\
    k_1=0.63576~~~~~k_2=1.87320~~~~~k_3=1.48695
\end{gathered}
\end{equation}
\!\!We used the following intuition to obtain this formula.
The negative KL-divergence goes to a constant as $\log\alpha_{ij}$ goes to infinity, and tends to $0.5\log\alpha_{ij}$ as $\log\alpha_{ij}$ goes to minus infinity.
We model this behaviour with $-0.5\log(1+\alpha_{ij}^{-1})$.
We found that the remainder $-D_{KL}+0.5\log(1+\alpha_{ij}^{-1})$ looks very similar to a sigmoid function of $\log\alpha_{ij}$, so we fit its linear transformation $k_1\sigma(k_2+k_3\log\alpha_{ij})$ to this curve.
We observe that this approximation is extremely accurate (less than $0.009$ maximum absolute deviation on the full range of $\log\alpha_{ij}\in(-\infty, +\infty)$; the original approximation \cite{kingma2015vdo} has $0.04$ maximum absolute deviation with $\log\alpha_{ij}\in(-\infty, 0]$).

One should notice that as $\alpha$ approaches infinity, the KL-divergence approaches a constant.
As in this model the KL-divergence is defined up to an additive constant, it is convenient to choose $C=-k_1$ so that the KL-divergence goes to zero when $\alpha$ goes to infinity.
It allows us to compare values of $\mathcal{L}^{\text{\tiny \emph{SGVB}}}$ for neural networks of different sizes.

\subsection{Sparsity}
From the Fig.~\ref{fig:kl} one can see that $-D_{KL}$ term increases with the growth of $\alpha$.
It means that this regularization term favors large values of $\alpha$.

The case of $\alpha_{ij}\to \infty$ corresponds to a Binary Dropout rate $p_{ij}\to 1$ (recall $\alpha=\frac{p}{1-p}$).
Intuitively it means that the corresponding weight is almost always dropped from the model.
Therefore its value does not influence the model during the training phase and is put to zero during the testing phase.

We can also look at this situation from another angle.
Infinitely large $\alpha_{ij}$ corresponds to infinitely large multiplicative noise in $w_{ij}$.
It means that the value of this weight will be completely random and its magnitude will be unbounded.
It will corrupt the model prediction and decrease the expected log likelihood. 
Therefore it is beneficial to put the corresponding weight $\theta_{ij}$ to zero in such a way that $\alpha_{ij}\theta_{ij}^2$ goes to zero as well.
It means that $q(w_{ij}\cond\theta_{ij}, \alpha_{ij})$ is effectively a delta function, centered at zero $\delta(w_{ij})$.
\begin{equation}
\begin{gathered}
    \theta_{ij}\to 0,\ \ \ \  \alpha_{ij}\theta_{ij}^2\to0\\
    \Downarrow\\
    q(w_{ij}\cond\theta_{ij},\alpha_{ij})\to\Normal(w_{ij}\cond0, 0)=\delta(w_{ij})
\end{gathered}
\end{equation}
In the case of linear regression this fact can be shown analytically.
We denote a data matrix as $X^{N\times D}$ and  $\alpha,\theta\in\RR^D$.
If $\alpha$ is fixed, the optimal value of $\theta$ can also be obtained in a closed form.
\begin{equation}
\label{eq:vdrvr}
\theta = (X^\top X + \diag(X^\top X)\diag(\alpha))^{-1} X^\top y
\end{equation}
Assume that $(X^\top X)_{ii} \neq 0$, so that $i$-th feature is not a constant zero.
Then from (\ref{eq:vdrvr}) it follows that $\theta_{i}=\Theta(\alpha_{i}^{-1})$ when $\alpha_{i}\to+\infty$, so both $\theta_{i}$ and $\alpha_{i}\theta_{i}^2$ tend to $0$.

\subsection{Sparse Variational Dropout for 
Fully-Connected and Convolutional Layers}
\label{sec:formulae}

Finally we optimize the \emph{stochastic gradient variational lower bound} (\ref{eq:sgvlb}) with our approximation of KL-divergence (\ref{eq:KL}).
We apply Sparse Variational Dropout to both convolutional and fully-connected layers.
To reduce the variance of $\mathcal{L}^{\text{\tiny \emph{SGVB}}}$ we use a combination of the Local Reparameterization Trick and Additive Noise Reparameterization.
In order to improve convergence, optimization is performed w.r.t. $(\theta, \log \sigma^2)$.
 
For a fully connected layer we use the same notation as in Section~\ref{sec:vdo}.
In this case, Sparse Variational Dropout with the Local Reparameterization Trick and Additive Noise Reparameterization can be computed as follows:
\begin{equation}
\label{eq:fc_vdo}
\begin{gathered}
%   b_{ij} \thicksim \Normal(A_i\theta_j, A_i^2(\pmb{\alpha}_j \odot \theta_j^2)) 
    b_{mj}\thicksim \Normal(\gamma_{mj}, \delta_{mj})\\
    \gamma_{mj} = \sum_{i=1}^Ia_{mi}\theta_{ij},\ \ \ \ \ 
    \delta_{mj} = \sum_{i=1}^Ia_{mi}^2\sigma_{ij}^2\\
\end{gathered}
\end{equation}
Now consider a convolutional layer.
Take a single input tensor $A_m^{H\times W\times C}$, a single filter $w_k^{h\times w\times C}$ and corresponding output matrix $b_{mk}^{H'\times W'}$.
This filter has corresponding variational parameters $\theta_k^{h\times w\times C}$ and $\sigma_k^{h\times w\times C}$.
Note that in this case $A_m$, $\theta_k$ and $\sigma_k$ are tensors.
Because of linearity of convolutional layers, it is possible to apply the Local Reparameterization Trick.
Sparse Variational Dropout for convolutional layers then can be expressed in a way, similar to (\ref{eq:fc_vdo}).
Here we use \label{fn2}$(\cdot)^2$ as an element-wise operation, $\ast$ denotes the convolution operation, $\text{vec}(\cdot)$ denotes reshaping of a matrix/tensor into a vector.
\begin{equation}
\begin{gathered}
    {\mathrm{vec}(b_{mk})} \thicksim \Normal(\gamma_{mk}, \delta_{mk})
\\
    \gamma_{mk} =\text{vec}(A_m\!\ast\!\theta_k),\ \ \ 
    \delta_{mk} = \diag(\text{vec}(A_m^2\!\ast\!\sigma_k^2))
\end{gathered}
\end{equation}
These formulae can be used for the implementation of Sparse Variational Dropout layers.
Lasagne and PyTorch source code of Sparse Variational Dropout layers is available at \url{https://goo.gl/2D4tFW}.
Both forward and backward passes through Sparse VD layers take twice as much time as passes through original layers.

\subsection{Relation to RVM}

% The only way to tune dropout rates during ordinary dropout training is to use a validation set or cross-validation.
% However, this approach is computationally intensive and is only practical for a small number of hyperparameters.

% On the other hand, Bayesian approach allows us to tune hyperparameters using optimization of a particular objective, known as evidence or marginal likelihood.
% This procedure is known as empirical Bayes or Evidence framework \cite{mackay1992bayes}.
% With the advances in stochastic variational inference, this procedure became scalable, which allowed one to tune a large amount of hyperparameters on a large dataset \cite{hoffman2013stochastic,kingma2013auto,challis2013gkl,titsias2014dsvi}.

The Relevance Vector Machine (RVM, \cite{tipping2001sparse}) is a classical example of a Sparse Bayesian model.
The RVM is essentially a Bayesian treatment of $L_2$-regularized linear or logistic regression, where each weight has a separate regularization parameter $\alpha_{i}$.
These parameters are tuned by empirical Bayes.
During training, a large portion of parameters $\alpha_i$ goes to infinity, and corresponding features are excluded from the model since those weights become zero.
This effect is known as Automatic Relevance Determination (ARD) and is a popular way to construct sparse Bayesian models.

Empirical Bayes is a somewhat counter-intuitive procedure since we optimize prior distribution w.r.t. the observed data.
Such trick has a risk of overfitting, and indeed it was reported in \cite{cawley2010overfit}.
However, in our work the ARD-effect is achieved by straightforward variational inference rather than by empirical Bayes.
Similarly to the RVM, in Sparse VD dropout rates $\alpha_i$ are responsible for the ARD-effect.
However, in Sparse VD $\alpha_i$ are parameters of the approximate posterior distribution rather than parameters of the prior distribution.
In our work, the prior distribution is fixed and does not have any parameters, and we tune $\alpha_i$ to obtain a more accurate approximation of the posterior distribution $p(w\cond \mathcal{D})$.
Therefore there is no risk of additional overfitting from model selection unlike the case of empirical Bayes.

That said, despite this difference, the analytical solution for maximum a posteriori estimation is very similar for the RVM-regression 
\begin{equation}
\begin{gathered}
\label{eq:rvm}
w^{MAP}=(X^\top X+\mathrm{diag}(\alpha))^{-1}X^\top y
\end{gathered}
\end{equation}
and Sparse Variational Dropout regression
\begin{equation}
\begin{gathered}
\label{eq:vdo}
\theta=(X^\top X + \diag(X^\top X)\diag(\alpha))^{-1}X^\top y
\end{gathered}
\end{equation}
Interestingly, the expression for Binary Dropout-re\-gu\-la\-ri\-zed linear regression is exactly the same as (\ref{eq:vdo}) if we substitute $\alpha_i$ with $\frac{p_i}{1-p_i}$ \cite{srivastava2014dropout}.

\section{Experiments}

\label{sec:exp}
We perform experiments on classification tasks and use different neural network architectures including architectures with a combination of batch normalization and dropout layers.
We explore the relevance determination performance of our algorithm as well as the classification accuracy of the resulting sparse model.
Our experiments show that Sparse Variational Dropout leads to extremely sparse models.

In order to make a Sparse Variational Dropout analog to an existing architecture, we only need to remove existing dropout layers and replace all dense and convolutional layers with their Sparse Variational Dropout counterparts as described in Section~\ref{sec:formulae} and use  $\mathcal{L}^{\text{\tiny \emph{SGVB}}}$ as the objective function.
The value of the variational lower bound can be used to choose among several local optima.

\subsection{General Empirical Observations}
We provide a general intuition about training of Sparse Bayesian DNNs using Sparse Variational Dropout.

As it is impossible for the weights to converge exactly to zero in a stochastic setting, we explicitly put weights with high corresponding dropout rates to 0 during testing.
In our experiments with neural networks, we use the value $\log\alpha=3$ as a threshold.
This value corresponds to a Binary Dropout rate $p>0.95$.
Unlike most other methods \cite{han2015learning, wen2016lssdnn}, this trick usually does not hurt the performance of our model.
It means that Sparse VD does not require finetuning after thresholding.

Training our model from a random initialization is troublesome, as a lot of weights become pruned away early during training, before they could possibly learn a useful representation of the data.
In this case we obtain a higher sparsity level, but also a high accuracy drop.
The same problem is reported by \cite{sonderby2016vae} and is a common problem for Bayesian DNNs.
One way to resolve this problem is to start from a pre-trained network.
This trick provides a fair sparsity level with almost no drop of accuracy.
Here by pre-training we mean training of the original architecture without Sparse Variational Dropout until full convergence.

Another way to approach this problem is to use warm-up, as described by \cite{sonderby2016vae}.
The idea is to rescale the KL-divergence term during training by a scalar term $\beta_t$, individual for each training epoch.
During the first epochs we used $\beta_t=0$, then increased $\beta_t$ linearly from 0 to 1 and after that used $\beta_t=1$.
The final objective function remains the same, but the optimization trajectory becomes different.
In some sense it is equivalent to choosing a better initial guess for the parameters.

We use the final value of the variational lower bound to choose the initialization strategy.
We observe that the initialization does not matter much on simple models like LeNets, but in the case of more complex models like VGG, the difference is significant.

On most architectures we observe that the number of epochs required for convergence from a random initialization is roughly the same as for the original network.
However, we only need to make a several epochs (10-30) in order for our method to converge from a pre-trained network.

We train all networks using Adam \cite{kingma2014adam}.
When we start from a random initialization, we train for 200 epochs and linearly decay the learning rate from $10^{-4}$ to zero.
When we start from a pre-trained model, we finetune for 10-30 epochs with learning rate $10^{-5}$.

\subsection{Variance Reduction}
To see how Additive Noise Reparameterization reduces the variance, we compare it with the original parameterization.
We used a fully-connected architecture with two layers with 1000 neurons each.
Both models were trained with identical random initializations and with the same learning rate, equal to $10^{-4}$.
We did not rescale the KL term during training.
It is interesting that the original version of Variational Dropout with our approximation of KL-divergence and with no restriction on alphas also provides a sparse solution.
However, our method has much better convergence rate and provides higher sparsity and a better value of the variational lower bound.

\begin{figure}[!t]
    \centering
    \includegraphics[width=0.38\textwidth]{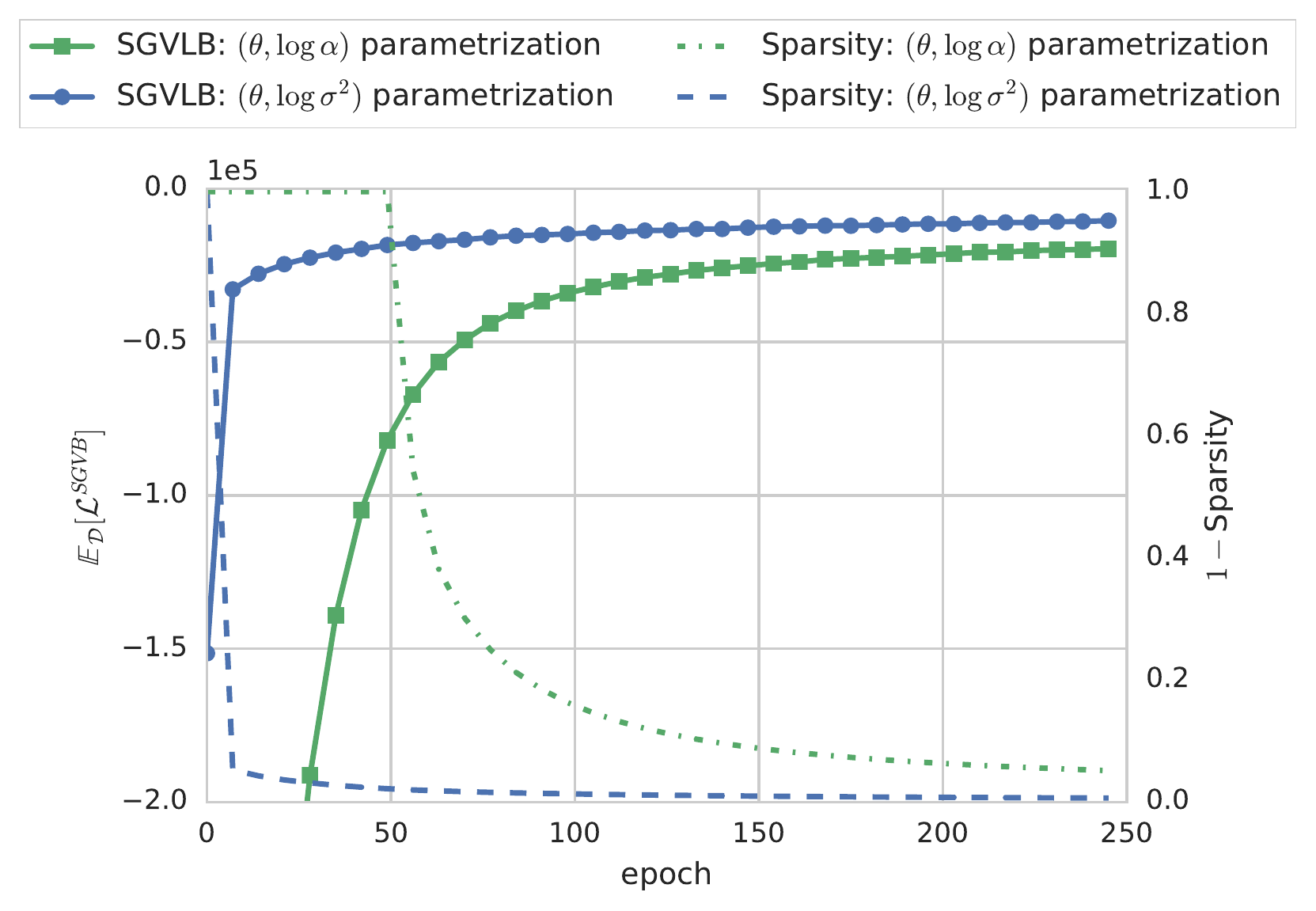}
    \caption{Original parameterization vs Additive Noise Reparameterization. Additive Noise Reparameterization leads to a much faster convergence, a better value of the variational lower bound and a higher sparsity level.}
    \label{pic:bpt_sgvlb}
\end{figure}

\subsection{LeNet-300-100 and LeNet5 on MNIST}
We compare our method with other methods of training sparse neural networks on the MNIST dataset using a fully-connected architecture LeNet-300-100 and a convolutional architecture LeNet-5-Caffe\footnote{A modified version of LeNet5 from \cite{lecun1998gradient}. Caffe Model specification: \url{https://goo.gl/4yI3dL}}.
These networks were trained from a random initialization and without data augmentation.
We consider pruning \cite{han2015learning,han2015deep}, Dynamic Network Surgery \cite{guo2016dynamic} and Soft Weight Sharing \cite{ullrich2017soft}.
In these architectures, our method achieves a state-of-the-art level of sparsity, while its accuracy is comparable to other methods.
It should be noted that we only consider the level of sparsity and not the final compression ratio.

 \begin{table}
 \small

  \label{table:lenet}
  \vskip 0.1in
  \centering
  \begin{threeparttable}
\resizebox{0.48\textwidth}{!}{\begin{tabular}{c@{\hspace{-0.1pt}}l@{\hskip -2pt}l@{\hspace{-2pt}}c@{\hskip -2pt}cc}
\hline
Network & Method&Error \% ~~~& Sparsity per Layer \% ~~~& $\frac{|\svec{W}|}{|\svec W_{\neq 0}|}$ \\ \hline
&\,Original
& ~~1.64 &  & 1 \\
&\,Pruning & ~~1.59 & $92.0 - 91.0 - 74.0$ ~~~~& 12 \\
\!\!\!\!LeNet-300-100~&\,DNS & ~~1.99 & $98.2 - 98.2 - 94.5$ ~~~~& 56 \\
& \,SWS & ~~1.94 &  & 23 \\
\multicolumn{1}{r}{\textcolor{gray}{(ours)}\!\!}&\,Sparse VD~~ & ~~1.92 & $98.9 - 97.2 - 62.0$~~~~ & \textbf{68} \\ \hline
&\,Original & ~~0.80 &  & 1 \\
&\,Pruning & ~~0.77 & $34 - 88 - 92.0 - 81$ ~~~~& 12 \\
\!\!\!\!LeNet-5-Caffe&\,DNS & ~~0.91 &$86 - 97 - 99.3 - 96$ ~~~~& 111 \\
&\,SWS & ~~0.97 & & 200 \\
\multicolumn{1}{r}{\textcolor{gray}{(ours)}\!\!}&\,Sparse VD~~ & ~~0.75 & $67 - 98 - 99.8 - 95$ ~~~~& \textbf{280} \\ \hline
\end{tabular}
}
  \end{threeparttable}
    \caption{Comparison of different sparsity-inducing techniques (Pruning \cite{han2015learning,han2015deep},
  DNS \cite{guo2016dynamic},
  SWS \cite{ullrich2017soft}) on LeNet architectures. Our method provides the highest level of sparsity with a similar accuracy.}
\end{table}

\subsection{VGG-like on CIFAR-10 and CIFAR-100}
To demonstrate that our method scales to large modern architectures, we apply it to a VGG-like network \cite{zagoruyko2015vgg} adapted for the CIFAR-10 \cite{krizhevsky2009learning} dataset.
The network consists of 13 convolutional and two fully-connected layers, each layer followed by pre-activation batch normalization and Binary Dropout.
We experiment with different sizes of this architecture by scaling the number of units in each network by $k\in\{0.25, 0.5, 1.0, 1.5\}$.
We use CIFAR-10 and CIFAR-100 for evaluation.
The reported error of this architecture on the CIFAR-10 dataset with $k=1$ is $7.55\%$.
As no pre-trained weights are available, we train our own network and achieve $7.3\%$ error.
Sparse VD also achieves $7.3\%$ error for $k=1$, but retains $48\times$ less weights.

\begin{figure*}[t!]
    \centering
    \begin{subfigure}{0.49\textwidth}
        \centering
        \includegraphics[width=0.68\textwidth]{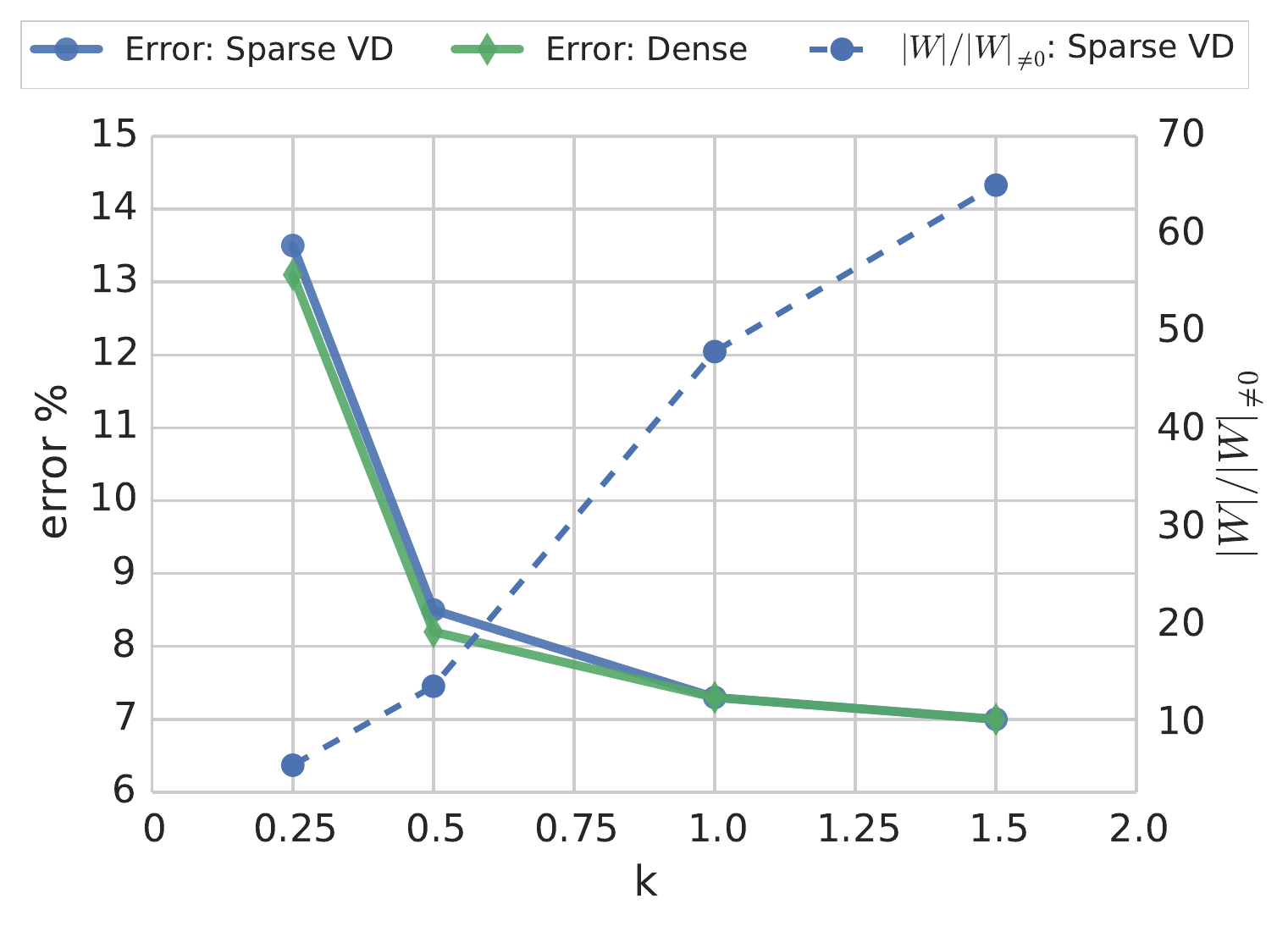}
        \caption{Results on the CIFAR-10 dataset}
    \end{subfigure}
    ~
    \begin{subfigure}{0.49\textwidth}
        \centering
        \includegraphics[width=0.68\textwidth]{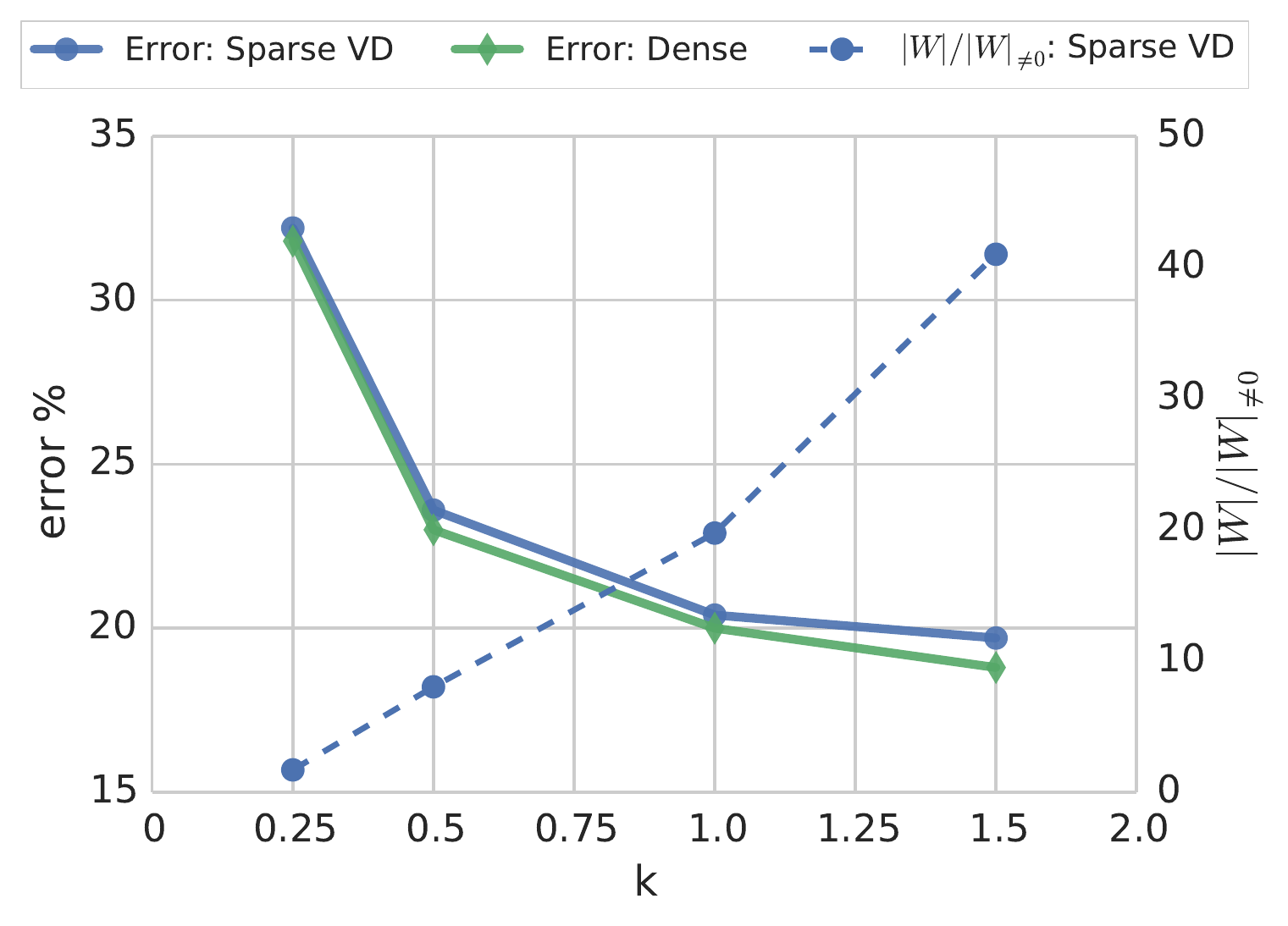}
        \caption{Results on the CIFAR-100 dataset}
    \end{subfigure}
    \caption{Accuracy and sparsity level for VGG-like architectures of different sizes. The number of neurons and filters scales as $k$. Dense networks were trained with Binary Dropout, and Sparse VD networks were trained with Sparse Variational Dropout on all layers. The overall sparsity level, achieved by our method, is reported as a dashed line. The accuracy drop is negligible in most cases, and the sparsity level is high, especially in larger networks.}\label{fig:animals}
\end{figure*}

We observe underfitting while training our model from a random initialization, so we pre-train the network with Binary Dropout and $L_2$ regularization.
It should be noted that most modern DNN compression techniques also can be applied only to pre-trained networks and work best with networks, trained with $L_2$ regularization \cite{han2015learning}.

Our method achieves over 65x sparsification on the CIFAR-10 dataset with no accuracy drop and up to 41x sparsification on CIFAR-100 with a moderate accuracy drop.

\subsection{Random Labels}

% \begin{table}
% \small
% \label{tab:random}
% \centering
% \caption{Experiments with random labeling. Sparse Variational Dropout (Sparse VD) removes all weights from the model and fails to overfit where Binary Dropout networks (BD) learn the random labeling perfectly.}
% \vskip 0.1in
% \label{table:rand}
% \resizebox{0.48\textwidth}{!}{\begin{tabular}{l|l@{\hskip -4pt}c@{\hskip -1pt}c@{\hskip -1pt}c}
% \hline
% Dataset & Architecture & Train acc.~~ & Test acc.~~ & Sparsity\\
% \hline
% MNIST & FC + BD & 1.0 & 0.1 & --- \\
% MNIST & FC + Sparse VD & 0.1 & 0.1  & 100\%\\
% % CIFAR-10 & VGG-like& 1.0 & 0.1 & --- \\
% CIFAR-10 & VGG-like + BD & 1.0 & 0.1 & --- \\
% CIFAR-10 & VGG-like + Sparse VD & 0.1 & 0.1 & 100\%\\
% \hline
% \end{tabular}}
% \end{table}

Recently is was shown that the CNNs are capable of memorizing the data even with random labeling \cite{zhang2016understanding}. The standard dropout as well as other regularization techniques did not prevent this behaviour.
Following that work, we also experiment with the random labeling of data.
We use a fully-connected network on the MNIST dataset and VGG-like networks on CIFAR-10.
We put Binary Dropout (BD) with dropout rate $p=0.5$ on all fully-connected layers of these networks.
We observe that these architectures can fit a random labeling even with Binary Dropout.
However, our model decides to drop every single weight and provide a constant prediction.
It is still possible to make our model learn random labeling by initializing it with a network, pre-trained on this random labeling, and then finetuning it.
However, the variational lower bound $\mathcal{L}(\theta, \alpha)$ in this case is lower than in the case of $100\%$ sparsity.
% Our results are presented in Table~\ref{table:rand}.
These observations may mean that Sparse VD implicitly penalizes memorization and favors generalization.
However, this still requires a more thorough investigation.

\section{Discussion}
The ``Occam's razor" principle states that unnecessarily complex should not be preferred to simpler ones \cite{mackay1992bayesian}.
Automatic Relevance Determination is effectively a Bayesian implementation of this principle that occurs in different cases.
Previously, it was mostly studied in the case of factorized Gaussian prior in linear models, Gaussian Processes, etc.
In the Relevance Tagging Machine model \cite{molchanov2015relevance} the same effect was achieved using Beta distributions as a prior.
Finally, in this work, the ARD-effect is reproduced in an entirely different setting.
We consider a fixed prior and train the model using variational inference.
In this case, the ARD effect is caused by the particular combination of the approximate posterior distribution family and prior distribution, and not by model selection.
This way we can abandon the empirical Bayes approach that is known to overfit \cite{cawley2010overfit}.

We observed that if we allow Variational Dropout to drop irrelevant weights automatically, it ends up cutting most of the model weights.
This result correlates with results of other works on training of sparse neural networks \cite{han2015deep,wen2016lssdnn, ullrich2017soft, soravit2017power}.
All these works can be viewed as a kind of regularization of neural networks, as they restrict the model complexity.
Further investigation of such redundancy may lead to an understanding of generalization properties of DNNs and explain the phenomenon, observed by \cite{zhang2016understanding}.
According to that paper, although modern DNNs generalize well in practice, they can also easily learn a random labeling of data.
Interestingly, it is not the case for our model, as a network with zero weights has a higher value of objective than a trained network.

In this paper we study only the level of sparsity and do not report the actual network compression.
However, our approach can be combined with other modern techniques of network compression, e.g. quantization and Huffman coding \cite{han2015deep,ullrich2017soft}, as they use sparsification as an intermediate step.
As our method provides a higher level of sparsity, we believe that it can improve these techniques even further.
Another possible direction for future research is to find a way to obtain structured sparsity using our framework.
As reported by \cite{wen2016lssdnn}, structured sparsity is crucial to acceleration of DNNs.

\section*{Acknowledgements} 

We would like to thank Michael Figurnov, Ekaterina Lobacheva and Max Welling for valuable feedback.
% Dmitry Molchanov was supported by the Laboratory of Tensor Networks and Deep Learning for Intellectual Data Analysis, Skolkovo Institute of Science and Technology (contract no. 14.756.31.0001 with Ministry of Education and Science of the Russian Federation),
Dmitry Molchanov was supported by the Ministry of Education and Science of the Russian Federation (grant 14.756.31.0001),
Arsenii Ashukha was supported by HSE International lab of Deep Learning and Bayesian Methods which is funded by the Russian Academic Excellence Project '5-100',
Dmitry Vetrov was supported by the Russian Science Foundation grant 17-11-01027.
We would also like to thank the Department of Algorithms and Theory of Programming, Faculty of Innovation and High Technology in Moscow Institute of Physics and Technology for the provided computational resources.

\bibliographystyle{icml2017}

\end{document}